\documentclass[sigconf]{acmart}
\usepackage{balance}
\usepackage{color}
\newif\ifworkinprogress
\workinprogresstrue
\ifworkinprogress
  \newcommand{\sm}[1]{\textcolor{blue}{\textbf{[Samaneh] #1}}}
  \newcommand{\mt}[1]{\textcolor{magenta}{\textbf{[Matthew] #1}}}
  \newcommand{\pv}[1]{\textcolor{green}{\textbf{[Puya] #1}}}
  
\else
  \newcommand{\sm}[1]{}
  \newcommand{\mt}[1]{}
  \newcommand{\pv}[1]{}
\fi

\graphicspath{{./img/}}

\usepackage{booktabs} 

\usepackage{subcaption} 
\usepackage{multirow} 
\usepackage{graphicx}
\usepackage{rotating}

\begin{document}
\title{Predicting Audio Advertisement Quality}

\copyrightyear{2018}
\acmYear{2018}
\setcopyright{acmcopyright}
\acmConference[WSDM 2018]{WSDM 2018: The Eleventh ACM International
Conference on Web Search and Data Mining }{February 5--9, 2018}{Marina
Del Rey, CA, USA}
\acmBooktitle{WSDM 2018: WSDM 2018: The Eleventh ACM International
Conference on Web Search and Data Mining , February 5--9, 2018, Marina Del
Rey, CA, USA}
\acmPrice{15.00}
\acmDOI{10.1145/3159652.3159701}
\acmISBN{978-1-4503-5581-0/18/02}

\author{Samaneh Ebrahimi}
\affiliation{%
  \institution{Georgia Institute of Technology}
  \streetaddress{}
  \postcode{}
}
\email{samaneh.ebrahimi@gatech.edu}

\author{Hossein Vahabi}
\affiliation{%
  \institution{Pandora Media Inc.}
  \streetaddress{}
  \postcode{}
}
\email{puya.vahabi@gmail.com}

\author{Matthew	Prockup}
\affiliation{%
  \institution{Pandora Media Inc.}
  \streetaddress{}
  }
\email{mprockup@pandora.com}

\author{Oriol	Nieto}
\affiliation{%
  \institution{Pandora Media Inc.}
  \streetaddress{}
  }
\email{onieto@pandora.com}

\renewcommand{\shortauthors}{Ebrahimi et al.}

%
\begin{abstract}
Online audio advertising is a particular form of advertising used abundantly in online music streaming services. In these platforms, which tend to host tens of thousands of unique audio advertisements (ads), providing high quality ads ensures a better user experience and results in longer user engagement. Therefore, the automatic assessment of these ads is an important step toward audio ads ranking and better audio ads creation.

In this paper we propose one way to measure the quality of the audio ads using a proxy metric called \emph{Long Click Rate (LCR)}, which is defined by the amount of time a user engages with the follow-up display ad (that is shown while the audio ad is playing) divided by the impressions. We later focus on predicting the audio ad quality using only acoustic features such as harmony, rhythm, and timbre of the audio, extracted from the raw waveform. We discuss how the characteristics of the sound can be connected to concepts such as the clarity of the audio ad message, its trustworthiness, etc.
Finally, we propose a new deep learning model for audio ad quality prediction, which outperforms the other discussed models trained on hand-crafted features. To the best of our knowledge, this is the first large-scale audio ad quality prediction study.

\end{abstract}

%


\maketitle
\section{Introduction}
\label{sec:introduction}
%

Audio advertising is abundantly used by online music streaming services. When users are listening to music, between two different songs,  they can be exposed to audio messages from advertisers. This form of ads are called audio ads and they can have a high impact on user listening experience. Given an opportunity space for ads, typically there is an ad retrieval and bidding process. After, the eligible ads are ordered based on the amount of money the advertisers are willing to pay times the quality of the ads. The challenge is how to measure the quality of the audio ads, and what are the key acoustic components of sound composition that create an engaging audio ads. 

Existing efforts toward discovering the quality of advertisements are mainly focused on visual perception: display ads and text ads \cite{Zhou2016PreClickQuality, LalmasFabrizio2015YahooGemini}. Therefore, the domain specific features considered in this context are text and image based features. Furthermore, in this case, there are well accepted proxy metric of ad quality,
such as clicks, dwell-time \cite{DwellTimeMicrosoft2014}, conversion or offensive rate \cite{Zhou2016PreClickQuality}.
However, in the context of audio ads it is not clear what kind of metric to consider as a proxy of quality as the audio ads can not be clicked. 
Therefore, the existing approaches are not directly applicable. Another challenge is extracting the features from sound content in order to build prediction models and understand how to create an engaging, high quality ads. 
To the best of our knowledge, there are no large scale audio ads quality studies that address these points.  

In this paper, we aim to predict the quality of ads using the acoustic signal.
The audio quality is not necessarily just the quality of the recording (clear vs. distorted), but may be implicit within the quality of an ad's overall composition. The composition of an ad includes the speaker tone and tempo, speaker gender, number of speakers, the message, music accompaniment, production effects and the overall mix of all of these sources. In the fields of \textit{Digital Signal Processing (DSP)} and \textit{Music Information Retrieval (Music-IR)} acoustic features are computed from the audio signal to capture elements of timbre, rhythm, and harmony, which can be used to model high-level subjective concepts (genre, mood) of sound and audio understanding \cite{muller2015fundamentals,sturm2014state}. 



For the quality metric we rely on a companion display ad that is presented at the same time the ad is playing. 
Users are listening to music and audio ads usually using their mobile device. 
While they are listening to audio ads they can decide to click on the associated banner during or after the audio ad is played. Following the click they are redirected to a landing page. The user will spend an amount of time in the landing page (dwell-time). If the dwell-time is higher than a threshold we consider the interaction to be engaging and we call it \emph{Long Click}. 
We use as a quality metrics \emph{Long Click Rate (LCR)} , i.e., the number of long click divided by the number of impressions. The main reasons behind not using the simple \emph{Click Through Rate} is that it can be very noisy because listeners may click on the banner by mistake, and because it may reflect only short-term user engagements ~\cite{Zhou2016PreClickQuality}. 

After defining the metric (\emph{LCR}), we propose a first acoustic based prediction model with handcrafted features. The main objective of this model is to have a prediction model that is interpretable, and that can tell us how to create a high quality audio ad. To improve the interpretation we also do a large scale user study. The results in this case shows that  speech with a moderate-tempo, articulate pronunciation, and moderate yet necessary vocal animation lead to better quality ads.
Finally, we also propose a deep learning model with the objective of improving the effectiveness of our prediction model, reaching an AUC of $0.79$ for cold-start audio ad quality prediction using only the spectrogram of audio.  

Let us resume the original contributions of this paper, where we:
\begin{itemize}
	\item Defined a metric to measure the quality of audio ads by relying on \emph{Long Click Rate (LCR) } on a companion display ad that appears at the same time the audio ad is playing.
	\item Conducted a large crowd-sourcing study to understand why an audio ad is more engaging than the other.
	\item Designed and implemented acoustic features to capture intuitive sonic attributes of timbre, rhythm, harmonic organization to build a prediction model for advertisement quality.
	\item Gave an interpretation on how low level acoustic features are connected to concepts such as speech speed, pronunciation, foreground (speech) to background sound (music and sound effects).
	\item Proposed different audio prediction models using only the acoustic features, reaching an AUC of $0.73$. 
	\item Proposed a deep learning model based directly on audio ad spectrograms, without using any hand-crafted features,, reaching an AUC of $0.79$.
\end{itemize}

\section{Related work}
\label{sec:related}
%
%

Computational advertising address the problem of ranking ads based on their relevance and quality. Recently, many different approaches have arisen to solve this problem, but they are all based on visual perception: display and text ads \cite{RosalesCIKM2012DISPLAY}, native and sponsored search \cite{SodomkaWWW2013SPONSOREDS}, and much more \cite{LalmasFabrizio2015YahooGemini}. Furthermore, they are all based on well-known metrics such as clicks, offensive rate \cite{Zhou2016PreClickQuality}, and dwell-time \cite{DwellTimeMicrosoft2014}. Our contributions are proposing a new metric for the quality of \emph{audio ads} that are not directly clickable, and a prediction model based only on acoustic signals. Next, we report the related work on classical ad quality prediction using visual perception, machine and deep learning algorithms, audio signals and music information retrieval.   

\noindent \textbf{Ad Quality.}
There are many existing approaches related to relevance of ads within sponsored search that are trying to optimize based on CTR ~\cite{ImprovingAdRelevance2013}, dwell-time ~\cite{DwellTimeMicrosoft2014}, revenue ~\cite{WangRevenue2015}, conversion rate ~\cite{KDD2012CONVERSION}. 
Recently, Zhou et al. ~\cite{Zhou2016PreClickQuality} focused on explicitly taking user ad feedback (offensive feed rate) into account to estimate ad quality.
Barbieri et al. ~\cite{Barbieri2016} presented a model on predicting the post-click user engagement using dwell-time for mobile ads.   
Chen ~\cite{chen2016deep} proposed a deep neural network (DNN) based model that directly predicts the CTR of an image ad based on raw image pixels and other basic features in one step. ~\cite{zhang2016deep} combines deep neural networks with factorization machine and also brings an improvement.

Most of the previous predictive models on ad quality are based on viewable features: display, native, and sponsored search ads ~\cite{ImprovingAdRelevance2013}. To the best of our knowledge, there is no work until now using the audio content directly to predict the user ad engagement after listening to ad.

\noindent \textbf{Audio Signal Understanding and Music-IR.}
Many tasks in music understanding work directly with the audio signals through the computation of hand-crafted, targeted features. 
These features include descriptions of \emph{timbre} to capture characteristics of the overall quality of the sound, \emph{rhythm} to capture relationships of sound event timing, and \emph{harmony} to capture relationships and patterns of harmonic frequency organization \cite{muller2015fundamentals,sturm2014state}. 
Rhythm features can be used to model tempo, beat locations, and meter. Harmony and melody features are used to model the notes present, the key, the mode, chords, and song structure. Timbre features can be used to model instrumentation and performer expression \cite{ellis2007beat,muller2015fundamentals,prockup2015modelingrhythm, nieto2014music}. Audio features of many types used in combination have also proved to be useful in modeling more nebulous concepts as well. Audio feature similarity can be used to derive music similarity and drive automatic play-listing as well as song search and retrieval \cite{knees2016introduction}. Supervised models using audio features can also be trained to capture very subjective concepts such as musical genre and taxonomy organization \cite{tzanetakis2002musical,prockup2015modelinggenre,sturm2014state}, assessing music and performer expression and sound quality \cite{bandiera2016good,wilson2016perception}, and even the mood of the music \cite{kim2010music}. Finally, Oord et al. proposed a deep learning approach for music recommendation \cite{van2013deep}.

In this work, we try to model the subjective concept of ad quality. Audio recording/sound quality is an important aspect of ad quality, and a well studied aspect in Music-IR (i.e., acoustical mixtures) and DSP in general (i.e., speech enhancement, synthesis) \cite{reiss2011intelligent,quatieri2002discrete}. However, audio ad quality refers to the quality of ad's compositional elements (message, speakers, music, sound effect). In the music domain, an analog to this work would be a quality assessment of the song and its musical composition. While it seems that is a difficult and subjective task for music, we feel that ad composition quality is a more tractable problem because ads are much shorter than most songs and are constructed to communicate a single, focused message.

\section{Problem Definition \& Preliminary Analysis}
\label{sec:problem}
%
\label{prob:def}
Defining a good or bad ad is not a trivial task. 
Let $A = {a1, a2, . . . , an}$ be a set ads, where each ad has a spectrogram of audio associated to it, and a set of acoustic features, our objective is to learn a function $f : A \rightarrow R_{\ge 0}$ that indicates the quality of an ad.
In order to learn a model for the audio ad quality prediction, first of all we need to define how to measure the audio ad quality, which are by their nature not clickable. 
After solving the measurement problem, we need to define the acoustic features in order to learn the quality prediction model. As a final step we need to define the learning model.


\subsection{Quality Measurement}
\label{prob:metrics}
\begin{figure}[t]
\includegraphics[width=.75\linewidth]{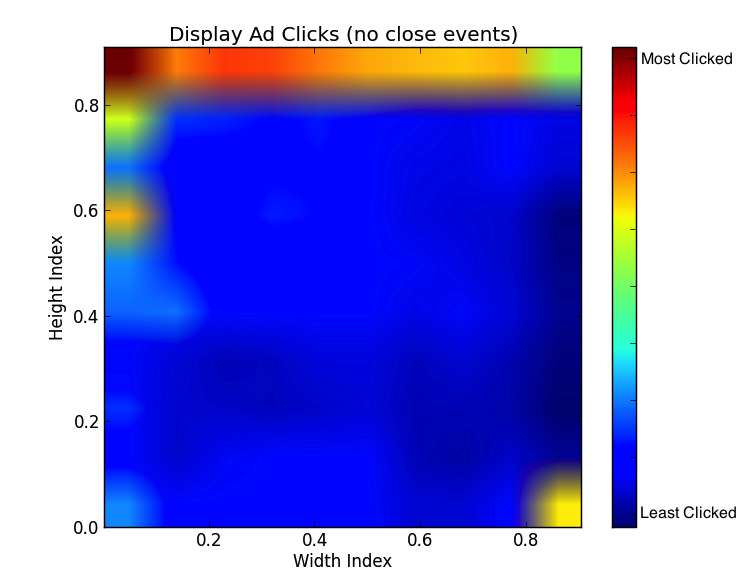}
\caption{Heat map of the clicks (for a selected device and screen size) when a display ad of full screen size dimension is shown. Red color indicates higher number of clicks. Most of the clicks appear to be near the close button (on the top left corner). These are most likely noisy clicks.}
\label{fig:heat_map}
\vspace{-0.3cm}
\end{figure}

One important step for predicting the quality of audio ad is to come up with useful quality metrics. So the major  question is how one can measure the quality of an audio ad. Unfortunately, in most platforms there is no access to explicit feedback from users. However, we can gather different implicit feedback from user performance. In this case, we rely on a companion display ad that is presented at the same time the audio ad is playing. When the audio is playing the user can decide to click on the companion banner. However the clicks can be very noisy as the user might just want to close the ad or just unintentionally click. Figure \ref{fig:heat_map} depicts a heat map of clicks on a mobile device. Most of the clicks are around the close button, and therefore most likely they are noisy clicks.

We use a new metric called \emph{Long Click Rate (LCR)}, i.e., the amount of times a user clicks on the companion banner, and stay in the follow up page for more than a threshold time \footnote{The effectiveness of the proposed metrics is arguable, as it is the validity of CTR. However dwell-time higher than a threshold is used successfully in literature as a user engagement metric in sponsored search \cite{DwellTimeMicrosoft2014} and native ads case \cite{LalmasFabrizio2015YahooGemini}. 
There might be a bias toward the images shown.} (and its variants). In this case we discovered that more than $90\%$ of the clicks on the top left corner clicked were resulting to user leaving in less than $5$ seconds the following page. Therefore we fixed the threshold to $5$ seconds. 

\begin{definition}
Given an ad $a \in A$, the total number of clicks with dwell-time higher than a threshold, ${N_{lc}}_a$, and the total number of times the audio ad was played, ${N_{}}_a$, we define \emph{Long Click Rate (LCR)} as
$LCR_a=(\frac{{N_{lc}}_a}{{N_{}}_a})$.
\end{definition}

We propose also a variant of the previous metric where we consider the vote for each user only once. 

\begin{definition}
Given an ad $a \in A$, the total number of unique users with a long click on ad $a$, ${N_{ulc}}_a$, and the number of users the audio ad was played to, ${N_{u}}_a$ , we define \emph{Robust Long Click Rate (R-LCR)} as
$R-LCR_a=(\frac{{N_{ulc}}_a}{{N_{u}}_a})$.
\end{definition}

Note that our objective is not predicting LCR or R-LCR, but we want to use them as proxy to find out whether an ad is ``good'' or ``bad''. In this sense we sort the ads by these metrics, and label as ``good'' the top $k$ percentile and ``bad'' the lower $k$ percentile. Details will be given on \ref{sec:dataset}.

\subsection{User Preferences}
\label{sec:userstudy}
We develop a user study to understand audio ad quality from users' point of view. This study can help us gain insight into the factors that affect user preferences for audio ads to inform feature engineering and the interpretation of results.  
To ensure a good variety in terms of audio ads quality we collect a set of ads from different \emph{LCR} quantiles. We focus on top five categories: Retail/Mass Retail, Automotive, Education, Health Care, Financial/Insurance. The design of this study is similar to the user study in ~\cite{Zhou2016PreClickQuality}. We provide evaluators with different pairs of two audio ads at a time and ask them to pick the ad they prefer. Moreover, to exclude the effect of evaluators' personal interests, at each comparison we ensure that ads are from the same category.
After evaluators made the comparison, we ask them to specify the reason behind their choice from a set of given probable reasons. To define such reasons, there are several work in the literature ~\cite{zhou2016predicting,o2010development}. However, these studies focus on display and native advertising. 
Therefore, we define a set of reasons specific to audio ads. The reasons are \textit{Clarity of Message} (clear message), \textit{Clarity of Sound} (High Quality, with balance between background music and voice, good repetition), \textit{Trustworthiness} (Speech matching the music, music/speech matching the product, not screaming buy), \textit{Aesthetic} (Music background, Sound effect, Single speaker, conversation, Speakers genders), and \textit{Brand} (Known Brand).
We ensure that each pair is shown to an assessor at most once, and the order of audio ad played will be randomly chosen in each comparison. To ensure the quality of assessments, we present each pair to three people and collect three independent judgments. Moreover, we employ a gold standard check (with a known low quality audio ad created by us), and a redundancy check. 

To perform the study, we collected 24 audio ads for each category randomly sampled from 6 distinctive quantiles of \emph{LCR} distribution. Our experiment consists of  276 pairwise comparison in each category resulting in 1380 total pairs. Ensuring 3 assessments for each pair and adding the gold standard check and redundancy check, we initially collect 4968 comparisons from 414 users. 
Also, we ensure users' demographics have proper diversity. Users are 56\% female and 44\% male, with age groups of ``18-25" (20.28\%), ``25-35" (23.19\%), ``35-45" (25.36\%), ``45-55" (16.18\%), ``55-65" (11.83\%), and ``65+" (3.14\%). Among all users, 78.5\% have yearly income of less than 50k, and remaining have income of 50-100k.
After removing users who failed the quality checks, 286 users are verified and included in our analysis. In Table ~\ref{tab:user_reasons} the percentage of each reason selected is presented. The results show that for all categories, around 80\% of time, \emph{Audio Aesthetic} is selected as a reason with high influence. \emph{Message clarity} and \emph{sound quality} are the next top reasons for users to select a good ad. These results show that audio aesthetics and sound quality are important aspects for determining the quality of audio ads, and suggest that these should be included in the models to predict the audio ad quality. This motivates the later work in this paper, using acoustic features derived from the audio signal that are designed to specifically capture elements of audio aesthetic, advertisement composition, and sound quality.


\begin{table}[t]
\centering
\small
\setlength\tabcolsep{2.4pt} 
\begin{tabular}{cccccc}
\hline
					& \textbf{Message}  	& \textbf{Sound} 	& \textbf{Trust-}	& \textbf{Audio} 		&       \\
\textbf{Category}  & \textbf{Clarity} 		& \textbf{Quality}	& \textbf{Worthiness} 	& \textbf{Aesthetic}	& \textbf{Brand}       \\ \hline
All                 & $0.71$	& $0.69$   	& $0.57$ 	& $0.82$	& $0.41$ \\
Automotive          & $0.67$    & $0.74$   	& $0.52$ 	& $0.83$  	& $0.40$ \\
Health Care         & $0.68$    & $0.67$   	& $0.56$ 	& $0.80$    & $0.34$ \\
Education           & $0.78$    & $0.68$   	& $0.62$ 	& $0.82$    & $0.43$ \\
Finance/Insurance & $0.72$    & $0.69$    & $0.56$ 	& $0.82$    & $0.39$ \\
Retail/Mass Retail  & $0.69$    & $0.69$   	& $0.56$ 	& $0.81$    & $0.47$ \\ \hline
\end{tabular}
\caption{Users' selected reason for preferring a particular audio ad. Audio Aesthetics is selected as a reason of preference in $80\%$ of time (for all categories).
}
\label{tab:user_reasons}
\vspace{-0.3cm}
\end{table}

\section{Acoustic Features}
\label{sec:feature_extraction}
%

In this section, we outline a set of methods and features motivated by the user study conducted, related work \textit{Digital Signal Processing (DSP)}  and \textit{Music Information Retrieval (Music-IR)} to obtain characteristics of the \textit{timbre}, \textit{rhythm}, and \textit{harmony} directly from the audio signal. These are summarized in Table \ref{tab:acousticfeatures}. Many of these features are high-dimensional, but are designed to capture intuitive auditory concepts in their dimensions. Because ad-quality is nebulous problem, we chose to start with these high-dimensional representations and allow the model to choose what is important. This affords the ability to gain an intuition as to what sonic attributes of an ad lead to its quality rating.

Acoustic features are computed from many sequential, fixed-length windows of audio samples and their resulting \textit{Discrete Fourier Transform (DFT)}. The \textit{Short-time Fourier Transform (STFT)}  is many sequential DFTs computed in sliding widows across the signal. The STFT shows the magnitude of specific frequencies present at at each time step in the audio signal, and is the starting point for many hand-crafted acoustic features \cite{Oppenheim:1999:DSP:294797,muller2015fundamentals}.
There are a few fundamental trade-offs when computing the STFT. In this work, all audio ads start as 128kbs mp3 files. Timbre and rhythm features use a frame length of 2048 samples with a sequential overlap of 50\% and 87.5\% overlap respectively. Harmony features use a frame length of 16384 with a 50\% overlap. These parameter values are quite common in the supporting literature \cite{muller2015fundamentals}.

\begin{table*}[t]
\centering
\linespread{0.9} 
\rmfamily 
\small
\begin{tabular}{llll}
\hline
\textbf{Type}& \textbf{Name}& \textbf{Dims} & \textbf{Description} \\ \hline
			& TFD 			& 10 	& Temporal Features and Dynamics - Block summary of of each frame's RMS amplitude and ZCR \cite{muller2015fundamentals,seyerlehner2010fusing}\\
\bf Timbre	& MFCC 			& 460 	& Block MFCCs - block summary of a compact snapshot of spectral shape \cite{muller2015fundamentals,seyerlehner2010fusing}\\
			& Delta-MFCC 		& 460 	& Block Delta MFCCs - Block summary of the change in spectral shape \cite{muller2015fundamentals,seyerlehner2010fusing}\\
			& MSP 			& 320 	& Block Mel-Spectral Patterns - Block summary of patterns in the activation of mel-frequency bands \cite{seyerlehner2010fusing} \\ \hline
			& TEMPO 		& 2 	& A primary and secondary estimate of the beats per minute \cite{prockup2015modelingrhythm}. \\
			& TG\_LIN		& 500 	& Tempogram TG - A histogram-like feature that shows emphasis of rhythmic repetition vs a bpm \cite{prockup2015modelingrhythm}  \\
\bf Rhythm	& TGR\{B,T,H\} 	& 13x3 	& Multi-band TG Ratios. A compact version of the tempogram at simple ratios of a tempo estimate \cite{prockup2015modelingrhythm} \\
			&BPDIST\{B,T,H\}& 36x3 	& Multi-band Beat Profile - Captures an snapshot of emphasis within estimated beat/pulse positions \cite{prockup2015modelingrhythm} \\
			& Mellin 		& 512 	& Mellin Scale transform - rhythmic self-similarity among multiple time scales \cite{holzapfel2011scale,prockup2015modelingrhythm} \\ \hline
			& SIHPCP		& 9 	& Shift-Invariant chroma histogram - 2D FFT of histogram of pitch classes (A, A\#, B, etc.) present \cite{gomez2006tonal}\\
\bf Harmony	& MODE 			& 1 	& Estimate of major or minor mode \cite{gomez2006tonal}\\
            & SICH,SICHC,SIKC&18x3 	& Shift-invariant (2D-FFT) histogram of chords present and the correlations of HPCP to chord and key templates \cite{gomez2006tonal} \\ \hline
\end{tabular}
\caption{Audio Acoustic Features Overview}
\label{tab:acousticfeatures}
\vspace{-0.3cm}
\end{table*}

\subsection{Timbre Features}
\textit{Timbre features} capture characteristics of the overall quality of the sound. Differences in timbre can distinguish male vs. female voice, multiple speakers, music vs. speech, sound effects, etc. Many of the timbre features are computed on each spectral frame, so in order capture a compact representation for the entire signal, they are summarized with their \textit{block statistics}. The mean and covariance (MCV) of the feature dimensions are computed on a set of fixed-length blocks of frames. The mean, variance, and the top right of the covariance matrix is vectorized to get an MCV vector for each block. All blocks are then summarized with the mean and variance of these MCV vectors \cite{seyerlehner2010fusing}. This summarization is performed for TFD, SSD, MFCC, and DMFCC features.

\textit{Temporal Features and Dynamics (TFD)} is a vector of the RMS energy and zero-Crossing rate (ZCR) of each audio frame. This captures a signals intensity, perceived volume, and noisiness.
\textit{Mel-frequency Cepstral Coefficients (MFCC)} are a compact representation of spectral shape motivated by how the human auditory system perceives sound. The frequency spectrum is warped with 128 perceptually motivated Mel filters. This perceptual frequency warping is linear up to 1kHz and logarithmic thereafter. Given a spectral frame $X$ and a Mel-spaced filter-bank matrix $F$, each Mel-spectrum frame is computed by $X'_m=X_k{F}$ for each mel-frequency $m$. The resulting Mel-Spectrum is then log scaled and the Discrete Cosine Transform (DCT) is performed to create the \textit{Ceptstrum} $C_c=\text{DCT}\{\log(X'_m)\}$ with Cepstral frequencies $c$. The MFCC feature is the first 20 coefficients of that Cepstrum, starting with the second coefficient ($c \in 2 \dots 21$). The first coefficient is often ignored because it represents scale rather than shape. This feature can be used to capture timbre, instrumentation, intonation, speaker/singer gender, number of speakers, tone, and the presence of music or sound effects. The \textit{Delta-MFCC} is the frame-to-frame difference of MFCCs, and captures the dynamics of these attributes \cite{muller2015fundamentals}.

\textit{Mel-Spectral Patters (MSP)} capture consistency, or temporal stability of the energy in the Mel-frequency bands. A coarse 32 band Mel-Spectrogram is computed from the STFT. A set of sequential 10-frame blocks is collected and activations in each band for each block are sorted. The feature is then summarized with the 60th percentile (median + 10\%) of all blocks in all bands \cite{seyerlehner2010fusing}.

\subsection{Rhythm Features}
\textit{Rhythm features} capture relationships of sound event timing. This can be used to model signal repetitiveness, how fast a speaker is talking, pulse/pace, and the presence of background rhythms if an the ad contains music backing tracks. Rhythm features start with the computation of an \textit{accent-signal}, which is a summarization of the STFT that shows emphasis of where new sonic events occur in time \cite{bock2013maximum}. This is also computed along three distinct frequency ranges to get sonic accents in low/bass (B: 27.5Hz-220Hz), middle/treble (T:220Hz-1.76kHz), and high (H:1.76kHz-14.08kHz) pitch accent signals \cite{prockup2015modelingrhythm}.
The \textit{Tempogram (TG\_LIN)} feature starts with an \emph{autocorrelation} of the accent signal $a$. The autocorrelation is a measure of self-similarity where $r[k]$ is the product of the two signals with respect to a time shift of lag $k$ (Eq. \ref{eq:acf}). 
\begin{equation}
\label{eq:acf}
r[k]=\sum_N a[n]a[n-k]
\end{equation}
The tempogram is the transformation of the lag axis (time-period) to a linearly-quantized frequency axis scaled to represent beats-per-minute. This captures absolute rates of repetitiveness in the accent signal, and is also used to estimate a primary and secondary \textit{Tempo Estimate (TEMPO)}. The TG\_LIN feature can be made more compact and tempo-invariant by looking at the weights in the tempogram at a set of simple fractional ratios $r$ of the estimated tempo $\tau$. This creates a \textit{Tempogram Ratio (TGR)} feature of relative temporal self-similarity (Eq. \ref{eq:tgr} \cite{prockup2015modelingrhythm}). 
\begin{equation}\label{eq:tgr}
\text{TGR}_r=\{4\tau,
\frac{8}{3}\tau,
3\tau,
2\tau,
\frac{4}{3}\tau,
\frac{3}{2}\tau,
\tau,
\frac{2}{3}\tau,
\frac{3}{4}\tau,
\frac{1}{2}\tau,
\frac{1}{3}\tau,
\frac{3}{8}\tau,
\frac{1}{4}\tau 
\}
\end{equation}

With a tempo estimate and the accent signal, musical \textit{beat-tracking} is performed via the dynamic programming method to capture positions of a fundamental pulse \cite{ellis2007beat}. The accent signal is summarized to create the \textit{Beat Profile (BP)}, which is the mean of the accent signal between successive beat positions \cite{prockup2015modelingrhythm}.


\subsection{Harmony Features}
\textit{Harmony features}, while seemingly tangential to this mostly non-music task can still provide important information regarding frequency organization. Speech is a very dynamic and harmonically rich signal. These features can capture a person's tone as monotone vs. animated and clearly pitched vs. raspy or breathy. If the ad has background music, they will capture the harmonic organization of that as well.

Harmony features start with a \textit{Constant-Q Transform (CQT)} performed on each STFT frame, which is a spectral filtering/warping to capture frequency emphasis at the locations of musical notes based on equal-tempered tuning. The \emph{Harmonic Pitch Class Profile (HPCP)} $P_{p,i}$ is a circular summarization of the Constant-Q Transform that captures the emphasis of the 12 pitch classes $p$ (A, A\#, B, etc.) for all octaves $k=0...K-1$ for each CQT frame $Q_{n,i}$ with note bins $n$ and frames $i$ (Eq. \ref{eq:chroma}). It can be further summarized by taking the mean across all frames. This can be made shift-invariant to key (harmonic tonic, anchor point) (SIHPCP) by taking the the Fourier Transform of the HPCP to emphasize the harmonic note relationships and density ]\cite{gomez2006tonal}.
\begin{equation}
\label{eq:chroma}
P_{p,i}=\sum_{k=0}^{K-1}Q_{12k, i} + p
\end{equation}

Harmony and Key specific correlation features are computed by finding the correlation of the HPCP to a set of contextual templates (chords, keys, etc.). The \textit{chordogram} is the correlation coefficients of the HPCP at each frame to a set of 12 major and 12 minor chords. The \textit{Chord Correlations (CHC)} feature is the mean of the chordogram. The \textit{Chord Histogram (CH)} is a histogram of chord estimates using the max of the chordogram at each frame. The \textit{Key Correlations (KC)} feature is the correlation coefficients of the mean of all frames of the HPCP to a set of 12 major and 12 minor key templates \cite{gomez2006tonal}. All of these are made shift-invariant (SICHC, SICH, SIKC) by taking the 2D-Fourier Transform (2D-FFT) of the resulting correlations/estimates. The feature becomes the Fourier Transform of the difference of the major and minor correlations/estimates concatenated to the sum of the major and minor correlations/estimates. Where {$\bf x$} is the concatenated major and minor template activations,
$${\bf x}_{maj}={\bf x}_{i=0\dots12} \hspace{.25in} {\bf x}_{min}={\bf x}_{i=13\dots24}$$
Where $X=\mathfrak{F}(x)$ is the 16-point Discrete Fourier Transform (DFT) of $x$,
$${\bf X}'_{sum}=\mathfrak{F}({\bf x}_{maj} + {\bf x}_{min}) \hspace{.25in} {\bf X}'_{diff}=\mathfrak{F}({\bf x}_{maj} - {\bf x}_{min})$$
$${\bf Y}_{shift-invariant} = \text{append}({\bf X}'_{sum},{\bf X}'_{diff})$$

\section{Modeling Using Acoustic Features}
\label{sec:model_MIR_feature_extraction}
%

In this section, we focus on predicting the quality of an audio ad, using acoustic features. We denote with $\nu \in R^d$ the feature vector, where $d$ is $2440$ dimension. Our goal is to predict the probability that an audio ad has high quality.

\subsection{Baseline Machine Learning Methods on Acoustic Features}
\label{sec:ML}

Due to curse of dimensionality, one important step is to reduce the number of variables or transform the features into a smaller dimension. There are total number of 2440 acoustic features extracted from audio ads which can have multi-collinearity or redundancy, and also can lead to over-fitting. We explore different dimension reduction and feature extraction techniques including PCA and Kernel PCA, feature selection based on feature importance calculated from Tree Classifiers, and L1 penalty on Logistic Regression and linear SVC. 

The first baseline classification method is \emph{Logistic Regression (LR)}. 
Denoting the quality as $y$ with binary values of 0 (bad quality), and 1 (good quality), the estimated probability of ad quality is:
\begin{equation}
\begin{aligned}
P(\hat{y} = 1| x)=\dfrac{1}{1+e^{-w^T x}}
\end{aligned}
\end{equation}
where $w \in d\times{1} $ is the vector to learn.
We use Logarithmic Loss (Logloss) in order to learn the previous function.  
Moreover, since we have many features, we use \emph{L1 regularized logistic regression} (L1-LR) which is often used for feature selection ~\cite{wainwright2007high,shevade2003simple}. In this model, analogous to Lasso~\cite{tibshirani1996regression}, penalty term is added to loss criterion, in other words we try to minimize the following function: 
\begin{equation}
L({w})=-\dfrac{1}{N}\sum_i{y_ilog{\hat{y}_i}+(1-y_i)log(1-\hat{y}_i)}+\lambda|{w}|
\end{equation}
Where $\lambda$ is the regularization parameter.
Furthermore, we implement  LR on the reduced dimension obtained with \emph{PCA} (LR-PCA),and \emph{Kernel PCA} (LR-KPCA).
Apart from the above mentioned methods, we explore the use of \emph{Support Vector Machines}, with linear support vector classifier (Linear-SVC) and nonlinear SVC with \textit{rbf} kernel (rbf-SVC). Furthermore, we use ensemble methods including \emph{Random Forest classifier} (RF)~\cite{breiman2001random}, Ada boost with Decision Trees (Ada-DT) and Bagging Classifier with Logistic Regression (BC-LR).

\subsection{Multilayer Perceptrons on Acoustic Features }
\label{sec:NN-Acoustic}
The Multilayer Perceptron (MLP)~\cite{witten2016data} is one of most commonly used neural network methods for classification. MLP is a feed-forward neural networks that use standard back-propagation algorithm for training. This classifier learns how to transform input data into a desired response in a supervise manner. The MLP consists of one input layer, one or more hidden layer and one output layer. 

\begin{figure}[t]
\includegraphics[width=\linewidth]{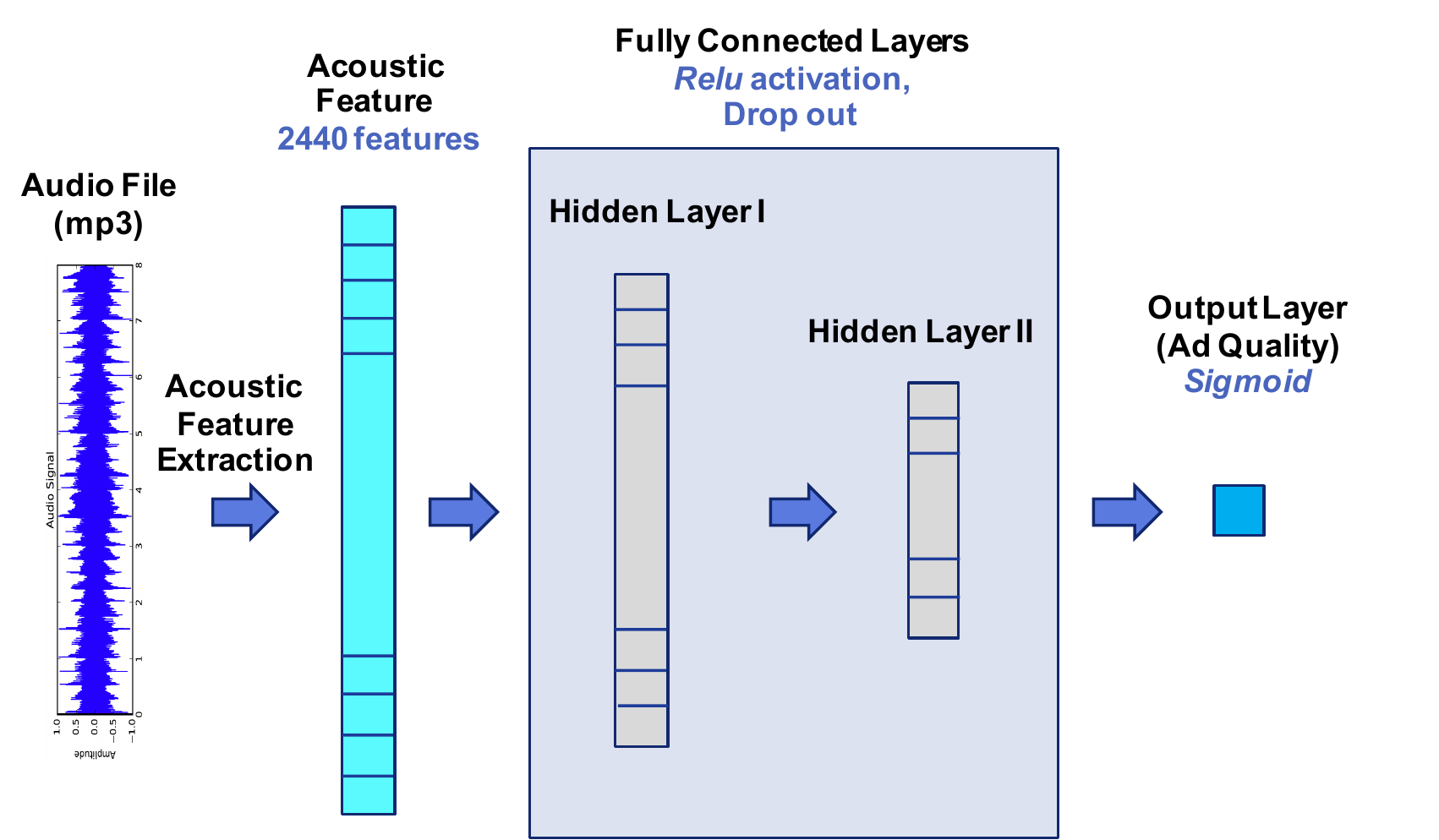}
\caption{MLP on Acoustic Features Architecture}
\label{fig:NN_acoustic_structure}
\vspace{-0.3cm}
\end{figure}

We implement MLP on audio ad's acoustic features (as the input layer), and train the model to gain the audio quality (as the output layer). Therefore, the architecture of the MLP method has 2440 neurons of input layer, and one neuron of the output layer (representing the predicted quality as good or bad). Hidden layers are fully connected layers with Rectified Linear Units (ReLUs) activation function, which is $f (x) = max(0, x)$. ReLUs ensure faster computation, more efficient gradient propagation, and sparse structure ~\cite{glorot2011deep}. Moreover, to prevent over-fitting and ensure regularization, Dropout is applied at the end of each layer~\cite{srivastava2014dropout}. The level of dropout, number of hidden layers, and number of neurons in each layer are tuned using cross validation in section~\ref{sec:tuning}. The output layer is one neuron with sigmoid activation function, representing the likelihood of good ad quality. 
The training process minimize the binary cross-entropy loss function and uses Adam as its optimizer with default parameters~\cite{kingma2014adam}. The MLP architecture with two hidden layer is shown in figure~\ref{fig:NN_acoustic_structure}. The presented MLP network can be used for predicting the audio ad's quality. Also, the last hidden layer can be seen as a new set of learned features from audio acoustic features. This features lie in a smaller dimension, and will be useful later for representing and clustering ads in a more informative way in direction of ads' quality.

\section{Deep Convolutional Networks on Spectrograms}
\label{sec:model_learnedCNN}
%

In the recent years, it is common in the field of music informatics to make use of Convolutional Neural Networks (CNNs) ~\cite{van2013deep} to estimate higher-level features directly from spectrograms.
These representations are typically contained in $\mathbb{R}^{F \times N}$ matrices with $F$ frequency bins and $N$ time frames, which become the input to the CNN. 

In this work, we propose a deep learning model to predict the quality of the audio ads using spectrogram. This model is inspired by music deep learning works ~\cite{van2013deep}.
We compute $F=100$ frequency bin, log-compressed constant-Q transforms  (CQT) \cite{schorkhuber2010constant} for all the ads in our dataset using \texttt{librosa} \cite{mcfee2015librosa} with these parameters: audio sampling rate at 44100Hz, hop length of 1024 samples, and 12 bins per octave. Furthermore, log-amplitude scaling is applied to the CQT spectrograms with a power law of 2 and scaling of 0.1. Even if CNNs allow for variable sized inputs due to the weight sharing nature of their convolutional filters, we need to normalize the sizes of all the spectrograms to train with minibatches of data points such that the gradients become more stable. To do so, and inspired by \cite{van2013deep}, we randomly sample three, 10-seconds long patches from each ad, resulting in a fixed-size $N$ input to the network, which are only used for training. We need patches because we need to place them in valid tensor for mini batch gradient during training. During inference the network is able to make predictions from full (i.e., not patched and variable $N$) audio ads.

\begin{figure}[t]
\includegraphics[width=\linewidth]{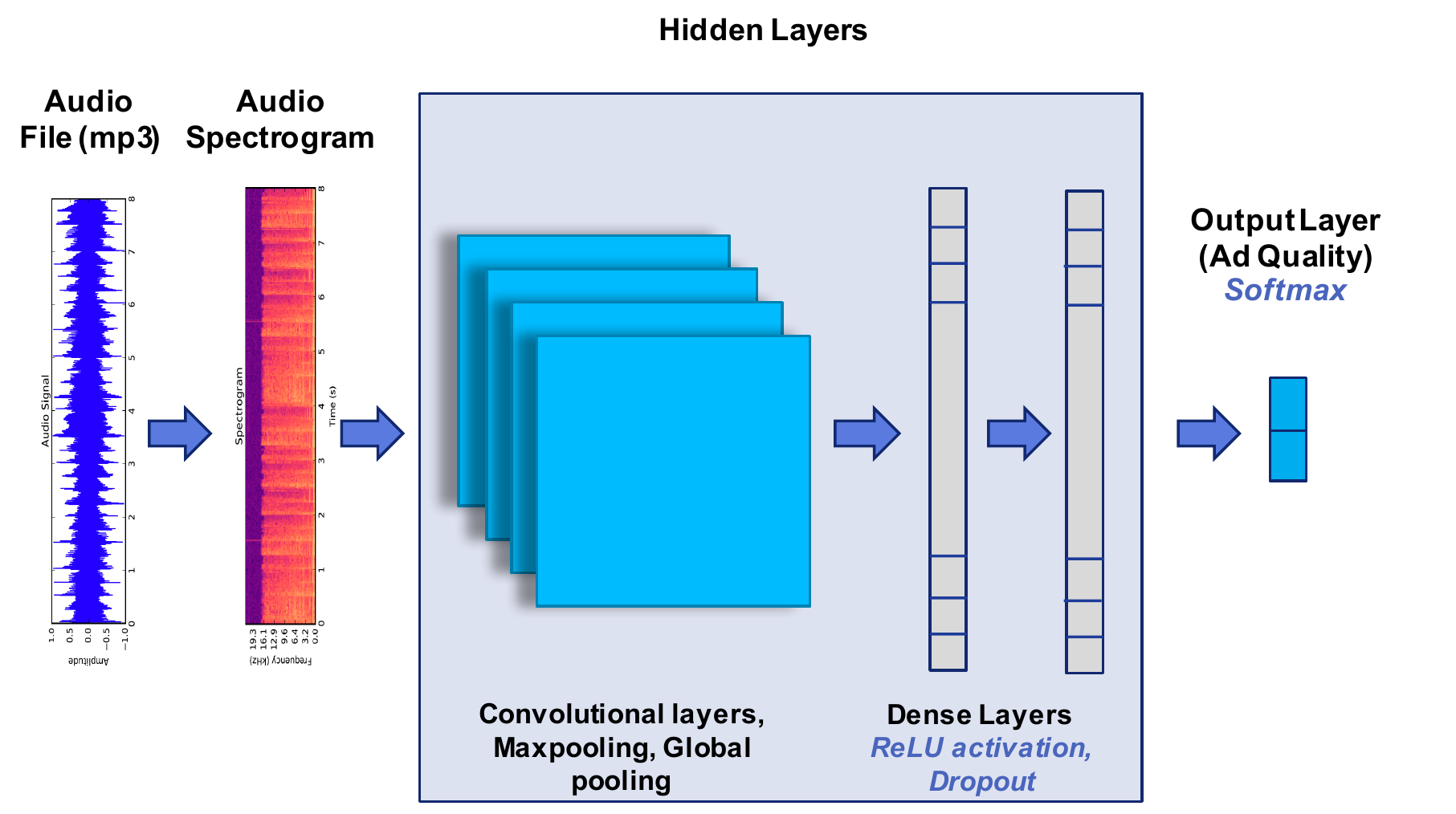}
\caption{CNN on Spectrograms Architecture }
\label{fig:CNN_spectrograms}
\vspace{-0.3cm}
\end{figure}

The network is composed of convolutional layers. All convolutions operate in the time axis only, thus having 1D convolutions instead of 2D. This is common practice when dealing with music signals \cite{van2013deep} and it is inspired by the importance of the relation between frequency bins, whose absolute placement in the spectrum matters. After these layers of convolution and maxpooling, a layer to compute global pooling through time is placed. This layer obtains 4 different statistics: mean, max, l2-norm, and standard deviation. Due to this aggregation, the network does not need to operate on fixed-sized spectrograms, therefore audio ads of different sizes will be able to be fed into the network. Finally, two dense layers with dropout and ReLUs are added after the global pooling, and the last layer is a set of two scalars with a softmax activation that represents the likelihood of each of the two classes: good or bad quality ad. The structure of our network is shown in figure
~\ref{fig:CNN_spectrograms}.
The training process minimizes the binary cross-entropy loss function, and uses Adam as its optimizer with the default parameters \cite{kingma2014adam}.

\section{Experiments}
\label{sec:experiments}
%
In this section we aim to evaluate our audio ad quality prediction models. We proceed by describing our dataset, parameter tuning, and we finish by discussing our results. 

\subsection{Datasets}
\label{sec:dataset}
We sample a set of audio ads from Pandora.com through May and June 2017. We then collected the information about the \emph{LCR} and \emph{R-LCR} for the audio ads following the procedure described in Section ~\ref{prob:metrics}.
For both labels we applied a filter of having at least two \emph{long clicks} or having at least two \emph{users long clicked} over $500$ impressions. The \emph{LCR}  related dataset  consists of 9k unique audio ads. The  
\emph{R-LCR} related dataset consists of 7k unique audio ads. The audio ads are in mp3 format with 128Kbps.

\begin{figure}[t]
  \centering
  \includegraphics[width=.85\linewidth]{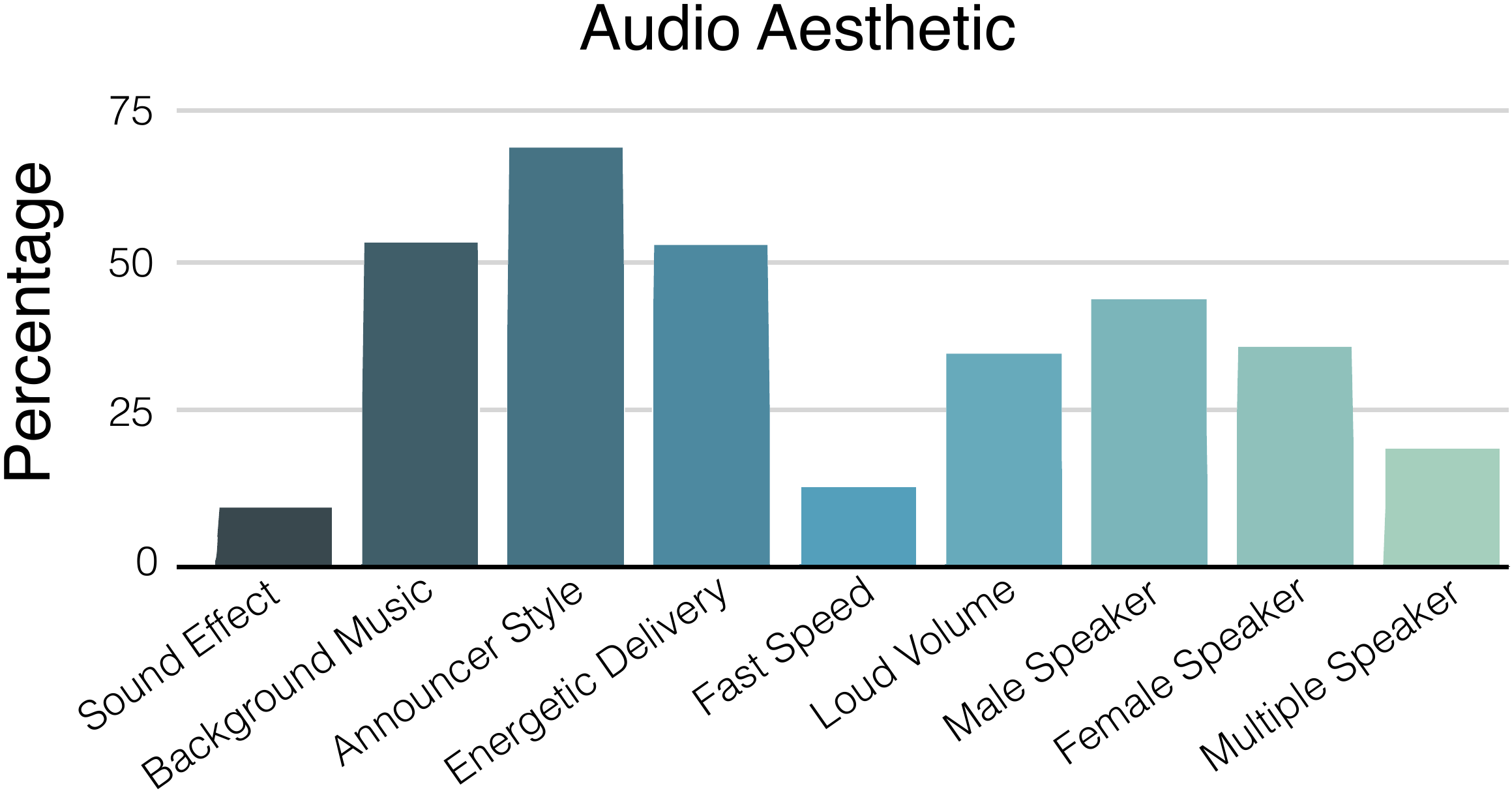}
  \caption{Percentage of Audio Aesthetics}
  \label{fig:aesthetic}
  \vspace{-0.3cm}
\end{figure}

In order to have some insight about our audio datasets, we sampled a set of $500$ audio ads, and asked human listeners to provide some information about their aesthetic data. The listeners determined the gender of speaker (female, male or more than one person), the speed of speaker, the existence of background music or sound effect in the ad, and the ad's volume. Figure~\ref{fig:aesthetic} show the percentage of aesthetic characteristics. Audio ads in our dataset seems to have a good diversity in terms of audio aesthetics. 
Our final dataset is made of one row per each audio ads with the acoustic features extracted and the audio spectrogram, and a binary label. In order to have a binary label based on \emph{LCR} and \emph{R-LCR}, we sort our dataset based on each quality metric and take the upper percentile as ``good'', and the lower percentile as ``bad''. To define which percentiles should be selected, we considered different scenarios and chose the percentiles with highest prediction accuracy. The scenarios we tried are [top $50$, lower $50$], [top $70$, lower $30$], [top $30$, lower $30$], and [top $10$, lower $10$] percentile as [``good'', ``bad'']. The best result was gained by labeling ``good'' the top $30$ percentile and ``bad'' the lower $30$ percentile.
 

%

\subsection{Implementation and Parameter tuning}
\label{sec:tuning}

To tune the parameters for each method and choose the best set of parameters, we implement 10-fold stratified cross validation on training data. In each fold, we use 9 fold to train the model and the remaining 1 fold for testing, and calculate the average test AUC. The average test AUC over all folds is used to select the best set of parameters.
For the MLP network explained in section~\ref{sec:NN-Acoustic}, batches of size 50 and total number 200 epochs are used for training, training takes around 110 seconds to finish on a 2.6 GHz Inter Core i7 processor with 16 GB 1600 MHz DDR3 memory. 
we implement the network with different dropout rate (None,20,40, and 50 percent), number of hidden layers (1,2,3,4), and number of neurons in each layer, and calculate average test AUC. The best configuration gained is dropout rate of 0.4 at the end of each layer, two hidden layer, and number of neurons of 150 and 75 in hidden layers I and II respectively.  
For CNN explained in section~\ref{sec:model_learnedCNN}, since we are sampling 3 patches from each audio in the dataset, we have a total of 32,406 patches, which we split for training and testing.
We make sure no patches from the same audio ad can be found in different splits (i.e., a given audio ad can not have patches both in, e.g., train and test).
Minibatches of 64 patches are used for training, and the network converges after 14 epochs, each of which takes around 6 minutes to finish on a Tesla M40 GPU. Once the network is trained, it takes 1.09 seconds in average to do inference on full ad spectrograms. 
After trying different parameters (including different number of convolutional layers), the best network is gained by using 4 convolutional layers (with 1024, 1024, 2048, 2048 filters, respectively) with length in time of the filters as 4, 4, 4, and 3, respectively. Maxpool layers are placed after each convolution, which aggregate through time with the following shapes: 8, 2, 2, 1. Also, two dense layers have 2048 neurons with 50\% dropout.

\subsection{Performance Analysis}
\label{sec:performance-comparison}
In this section, we compare the effectiveness of proposed methods with best configuration. For this comparison, the baseline methods are the top 5 machine learning approaches (in terms of test AUC) explained in Sec.~\ref{sec:ML}. The results are presented in table~\ref{tab:comparison_methods_AUC}.
As the results shows, among the baseline methods, logistic regression with L1 regularization gives the highest AUC for both labels. Moving to MLP network on acoustic features, the AUC increases by 7.4\% and 6.6\% for \emph{LCR}, and \emph{R-LCR} labels respectively. Furthermore, the CNN on audio spectrograms gives the highest accuracy of 0.7492 and 0.7906 test AUC for \emph{LCR} and \emph{R-LCR} labeled data. In this case, in comparison to the best baseline method, the accuracy increases by 18.19\% and 14.76\% for \emph{LCR} and \emph{R-LCR} labels. This results shows that CNN method on spectrograms beats the methods based on acoustic features, and audio spectrograms itself can be used as a sufficient input to obtain the quality of audio ad.

Another comparison, is based on the runtime of different methods. Test and training time (in seconds) for each method is recorded in Table~\ref{tab:Runtime}. The results show that CNN methods have much higher training time in comparison to other methods. Therefore, if the size of data increases, the run time can be hindering in comparison with other methods.

\begin{table}[t]
\centering
\begin{tabular}{llll}
\hline
\multicolumn{2}{c}{\multirow{2}{*}{\textbf{Method}}} & \multicolumn{2}{c}{\textbf{Dataset}} \\ \cline{3-4} 
\multicolumn{2}{c}{} & \textbf{Pandora LCR} & \textbf{Pandora R-LCR} \\ \hline
\multirow{5}{*}{ML} 
& LR & $0.5960(\pm0.01815)$ & $0.6582(\pm0.0175)$ \\
 & LR+L1 &  $\textbf{0.6338}(\pm0.0172)$& $\textbf{0.6889}(\pm0.0157)$  \\
 & rbf-SVC & $0.5801(\pm0.0195)$ & $0.6238(\pm0.0273)$\\
 &  Ada-DT & $0.5641(\pm0.0117)$ & $0.6105(\pm0.0186)$ \\
 &  RF & $0.579(\pm0.018)$  & $0.6080(\pm0.0233)$ \\
 \hline
\multirow{2}{*}{MLP} & MLP\_2layer & $\textbf{0.6808}(\pm0.0237$) &  $\textbf{0.7342}(\pm0.0146$) \\
 & MLP\_3layer &  ${0.6657}(\pm0.0173$) &  ${0.7299}(\pm0.0141$) \\ \hline
CNN & CNN\_4layer & $\textbf{0.7491} (\pm0.0224)$ & $\textbf{0.7906} (\pm0.0207)$ \\ \hline
\end{tabular}
\caption{Average test AUC (and its $95\%$ confidence interval)}
\label{tab:comparison_methods_AUC}
\vspace{-0.3cm}
\end{table}

\begin{table}[t]
\centering
\begin{tabular}{lll}
\hline
\textbf{Method} & \textbf{Training Time (sec)} & \textbf{Prediction Time (sec)} \\ \hline
LR & 12.4626 & 0.0017 \\
LR\_L1 &  118.1302& 0.0022 \\
rbf-SVC & 54.3700 & 6.0630 \\
Ada\_DT & 61.9137 & 0.0352 \\
RF & 1.9384 & 0.0143 \\ 
\hline
MLP\_2layer & 111.1795 & 0.0852 \\
MLP\_3layer & 115.4068 & 0.1274 \\
\hline
CNN\_4layers & 3,540.1200 & 1.0900 \\ \hline
\end{tabular}
\caption{Runtime comparison for different methods}
\label{tab:Runtime}
\vspace{-0.3cm}
\end{table}

\subsection{Intuition and Discussion}
\label{sec:intuition}

\begin{figure}[h]
\includegraphics[width=1.0\linewidth]{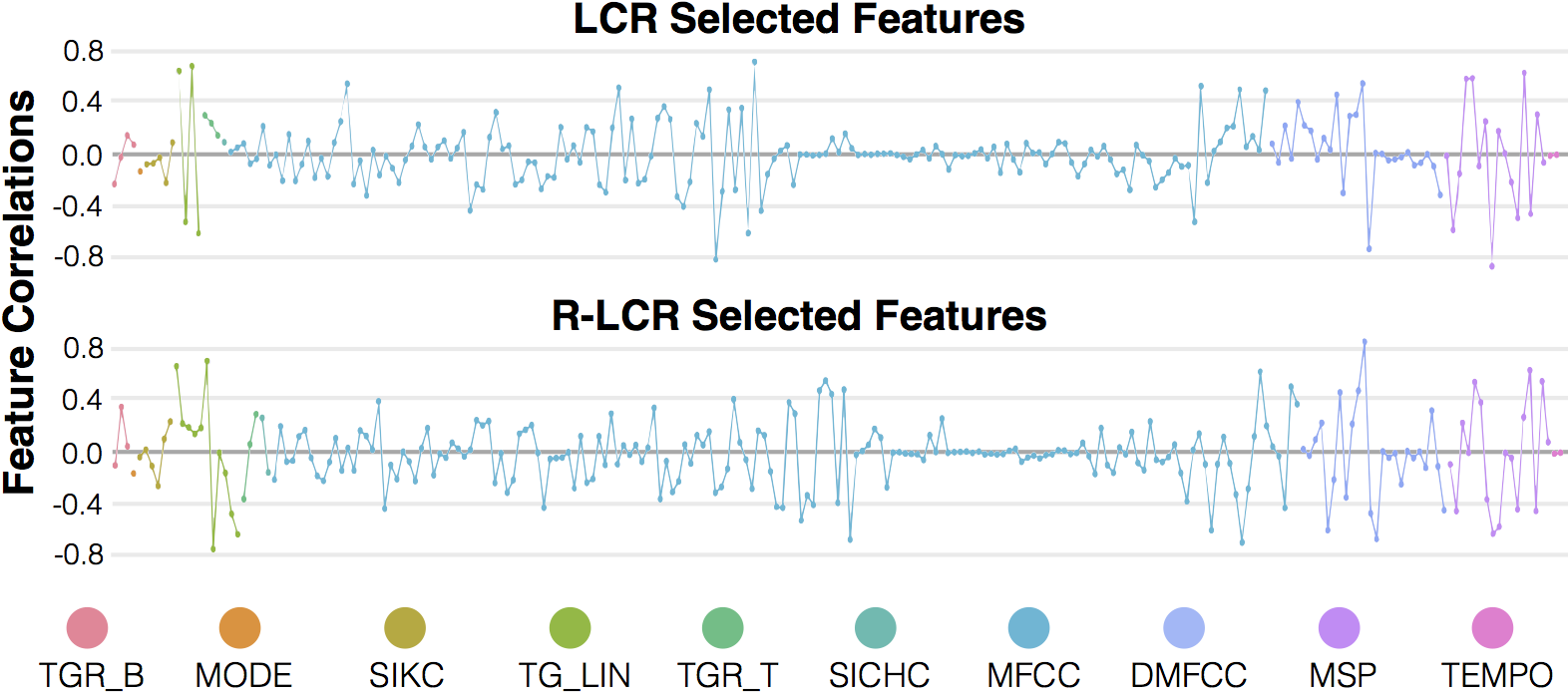}
\caption{Features selected by L1 regularization and their corresponding coefficient in Logistic Regression.}
\label{fig:L1_coeffs}
\vspace{-0.3cm}
\end{figure}

One good property of ML-based methods is their easy interpretation. Among ML methods, the best AUC accuracy was obtained using LR-L1 method. In this method, L1 regularization is used for variable selection. The list of selected variables and their corresponding coefficients are presented in Figure~\ref{fig:L1_coeffs}  for \emph{LCR} and \emph{R-LCR} data respectively. From the selected features and their value we can get insight about some general characteristics in favor of a better audio ad.   

\noindent\textbf{Acoustic Feature Intuition.}
The \textit{timbre features} were quite important when predicting ad-quality. Correlations in the Mel-Frequency Cepstral Coefficients (\emph{MFCC}) and Delta MFCC (\emph{DMFCC}), which correspond to the spectral shape of the speech and its variance, suggest that clear pronunciation is important. Furthermore, jarring sound effects and music can mask the parts of the spectrum that contain the speech, leading to poor quality. Correlations of Mel-Spectral Patterns (\emph{MSP}) dimensions suggest that if low frequencies or high frequencies are too active, the ad is of low quality, suggesting the balance of the audio mix is bad. Furthermore high variance in low and high frequencies leads to bad quality as well, as the overall spectral range of sound is not stable. Because speech sits in the middle frequencies, the overbalance and variance of frequencies surrounding speech can be distracting.

From the \textit{rhythm features}, specifically \emph{TGR} and \emph{TG\_LIN}, faster relative pulse repetition was negatively correlated with quality, and slower repetition relative to a pulse was positively correlated quality. This indicates that speaking more slowly and moderate tempo music (if any) correlates to better quality.
 
The \textit{harmony features} also showed some interesting correlations. Major \emph{MODE} positively correlates with quality, suggesting the choice of music sounds (if any) in a major key. \emph{SIKC} and \emph{SICH}, while having little direct contextual meaning in speech signals, capture patterns and transitions in the harmonic spacing of sounds. Animated speech, where pitch shifts up and down, should be captured. These correlations show that moderately animated pitch shifts lead to good quality ads. If speech is too animated (too much shift) or too monotone (too little shift) it leads to poor quality. This suggests to speak with a slightly animated tone and to choose music (if any) with a little harmonic motion.

\noindent\textbf{Qualitative Listening Analysis.}
Further intuition can be gained by listening to the ads that are correctly predicted as having good versus poor quality. This is done by clustering audio features as well as the last layer of the neural-network for each of the model types and listening to the ads of clusters that are primarily true-positives or true-negatives. Doing so, neural-network based methods gives much better clusters in terms of separating high  and low quality audio ads. By listening to the clusters with highest and lowest quality we can gain insights about differences between these clusters. This qualitative analysis showed that \emph{Good Ads} have clear, mid-paced, solo voice. They contain moderately varied, non-monotone speaker expression with moderate excitement. The speech is conversational between the speaker and the audience (i.e., story telling, call-to-action) or between two isolated speakers in the ad. If there is background music, it is basic. There is a good balance between background and foreground sounds, and there are almost no sound effects. \emph{Poor quality ads} have faster paced language, long winded explanations of products, and very monotone expression. Many of them had loud backing music and jarring sound effects that sometimes obscure the speech. Finally many of these were also lower quality recordings with distortion and compression effects. These qualitative attributes were shared among the models, but were more clear to interpret in the LR-L1 method than the neural network method. It further suggests that these acoustic attribute saliences exist, echoed both in the quantitative feature correlation analysis and this qualitative analysis. However, the neural network, which has Higher AUC overall, is potentially picking up on more subtleties, which leaves room for further exploration into the additional attributes it is capturing.





\section{Conclusions and future work}
\label{sec:conclusion}
%

In this work we presented the first audio ad quality prediction model. We focus on building a prediction model that is using only acoustic content. 
Given the fact that there is no direct feedback from users regarding audio ads, we proposed two quality metrics based on \emph{Long Click Rate (LCR)} on the audio ad's companion display for labeling the audio files. We than conducted a user study to understand the users' reasons for prefering a particular audio ad. The results of the user study show that in $80\%$ of the cases they prefer an audio ad due to audio aesthetics. Motivated by this, we implemented several acoustic features. We later used these features to learn a prediction model for audio ad quality. The results show that speaking slow with less jarring sound effect and only simple music (if any) can lead to better audio ad quality. Another finding is that conversational tone are usually more engaging, for instance \textit{"Do you have a need? ... try this"} is better than \textit{"We have the best deal down at car store."}. Finally we proposed a deep convolutional network on audio spectrogram, reaching an AUC of $0.79$ for the prediction task. 

For future works, we plan to study further the proposed quality metrics, for instance by investigating if there is a bias toward the images shown. Furthremore, we want to include in the model the user profile features, the campaign information and historical data of ad quality. Another interesting investigation is to use the transcription of the audio ads content and perform text analysis.

\balance
\bibliographystyle{ACM-Reference-Format}
\bibliography{bibliography} 
\end{document}